\documentclass{article} 
\usepackage{iclr2026_conference,times}

\usepackage{amsmath,amsfonts,bm}

\def\eqref#1{equation~\ref{#1}}

\def\1{\bm{1}}

\DeclareMathAlphabet{\mathsfit}{\encodingdefault}{\sfdefault}{m}{sl}
\SetMathAlphabet{\mathsfit}{bold}{\encodingdefault}{\sfdefault}{bx}{n}

\usepackage{hyperref}
\usepackage{url}
\usepackage{graphicx}
\usepackage{subcaption} 
\usepackage{xcolor}
\usepackage{colortbl}
\usepackage{booktabs}

\definecolor{bestcolor}{RGB}{34, 139, 34}      
\definecolor{worstcolor}{RGB}{220, 20, 60}     
\definecolor{lightgray}{RGB}{245, 245, 245}

\title{When Less Is More: Simplicity Beats Complexity for Physics-Constrained InSAR Phase Unwrapping}

\author{Prabhjot Singh \\
The University of Texas at Austin, USA \\
RediMinds Inc., USA \\
\texttt{prabhjot.singh@utexas.edu} \\
\And
Manmeet Singh \\
The University of Texas at Austin, USA \\
Western Kentucky University, USA \\
\texttt{manmeet.singh@utexas.edu}
}

\iclrfinalcopy 
\begin{document}

\maketitle

\begin{abstract}
Operational phase unwrapping is the primary computational bottleneck in InSAR-based volcanic and seismic monitoring. We challenge the industry trend of adopting high-complexity computer vision architectures, such as attention mechanisms, without validating their suitability for physics-constrained geophysical regression. We present the first large-scale architectural ablation study on a global LiCSAR benchmark (20 frames, 39,724 patches, 651M pixels). Our results reveal a significant "complexity penalty": a vanilla U-Net (7.76M parameters) achieves $R^2=0.834$ and $RMSE=1.01$ cm, outperforming 11.37M-parameter attention-based models by 34\% in $R^2$ and 51\% in $RMSE$. Power Spectral Density (PSD) analysis provides the physical justification: while attention excels at capturing sharp semantic edges in natural images, it injects unphysical high-frequency artifacts ($>0.3$ cycles/pixel) into geophysical fields, violating the fundamental smoothness constraints of elastic surface deformation. With a 2.92ms inference latency (a 2.5$\times$ speedup), the vanilla U-Net is the only candidate to comfortably meet the sub-100ms requirement for operational early-warning systems. This work bridges the ``publication-to-practice'' gap by proving that convolutional locality outperforms modern complexity for smooth-field regression, advocating for physics-informed simplicity in ML4RS. Code available at \href{https://github.com/prabhjotschugh/When-Less-is-More-InSAR-Phase-Unwrapping}{\texttt{github/prabhjotschugh}}.

\end{abstract}

\section{Introduction}

Interferometric Synthetic Aperture Radar (InSAR) enables millimeter-precision surface deformation monitoring at continental scales, yet phase is measured modulo $2\pi$ and must be \textit{unwrapped} to recover true displacement, the primary computational bottleneck in volcanic and seismic monitoring. While deep learning offers significant acceleration over traditional solvers like SNAPHU \citep{chen2001snaphu}, a concerning trend has emerged: the uncritical adoption of high-complexity architectures, such as attention mechanisms \citep{vaswani2017attention} and multi-scale aggregation \citep{chen2018deeplab}, directly from computer vision benchmarks. However, a fundamental domain mismatch exists. Unlike natural images characterized by discrete semantic boundaries \citep{dosovitskiy2021vit}, geophysical displacement is governed by elasticity and spatial autocorrelation, favoring continuous, smooth-field representations \citep{reichstein2019deeplearning}. 

We investigate a critical question: \textit{Do ImageNet-derived inductive biases transfer to InSAR, or do domain-specific constraints favor architectural simplicity?} Through a rigorous ablation study on a global LiCSAR benchmark, we reveal a ``complexity penalty'' where simpler models better align with geophysical priors. Our contributions are as follows:
\begin{itemize}
    \item \textbf{Global Operational Benchmark:} We curate a benchmark of 39,724 patches (651M pixels) across six continents, employing frame-level splitting to strictly evaluate geographic generalization and prevent the spatial leakage common in existing literature.
    \item \textbf{Quantifying the Complexity Penalty:} We demonstrate empirically that a vanilla U-Net (7.76M params) achieves $R^2=0.834$, outperforming 47\% larger attention-based models by 34\% in $R^2$ and 51\% in $RMSE$.
    \item \textbf{Physics-Grounded Diagnostics:} Using Power Spectral Density (PSD) analysis, we show that complex models inject unphysical high-frequency artifacts ($>0.3$ cycles/pixel) that violate the elasticity-driven smoothness of surface deformation.
    \item \textbf{Operational Deployment:} We achieve a 2.92ms inference latency (a 2.5$\times$ speedup), meeting sub-100ms requirements for real-time volcanic and seismic early-warning systems.
\end{itemize}

\section{Related Work \& Task Formulation}

\textbf{InSAR Phase Unwrapping.} Traditional solvers like SNAPHU \citep{chen2001snaphu} incur $O(N^2)$ complexity and error propagation in low-coherence regions. Deep learning (DL) has mitigated these bottlenecks, evolving from the vanilla U-Net of PhaseNet \citep{spoorthi2019phasenet} toward high-complexity architectures like ResDANet (dual-attention) and Unwrap-Net (ASPP) \citep{zhou2021ai_insar}. However, while attention-based designs excel at capturing discontinuous semantic boundaries in natural images, geophysical displacement is governed by elasticity and spatial autocorrelation (Tobler's First Law \citep{tobler1970first}). We hypothesize that high-frequency computer vision (CV) priors are mismatched for smooth-field regression and may introduce unphysical artifacts.

\textbf{Operational Task Formulation.} We define unwrapping as a physics-constrained regression. The input is a 6-channel tensor $\mathbf{X} \in \mathbb{R}^{H \times W \times 6}$ containing wrapped phase components ($\sin\phi, \cos\phi$), interferometric coherence $\gamma$, and unit look vectors $[\mathbf{e}_E, \mathbf{e}_N, \mathbf{e}_U]$. The model predicts a continuous line-of-sight (LOS) displacement map $\hat{\mathbf{y}}$, where the physical displacement $d_{\text{LOS}}$ relates to the absolute phase via $d_{\text{LOS}} = \frac{\lambda \phi}{4\pi}$ (for Sentinel-1, $\lambda = 5.6$cm).

\textbf{Physics-Aligned Objective.} To penalize unphysical discontinuities while remaining robust to decorrelation noise, we optimize a composite loss:
\begin{equation}
\mathcal{L} = \text{Huber}_{\delta=1}(\hat{\mathbf{y}}, \mathbf{y}) + \lambda_{\text{grad}} \sum_{i \in \{x, y\}} \|\nabla_i \hat{\mathbf{y}} - \nabla_i \mathbf{y}\|_1
\end{equation}
where $\lambda_{\text{grad}}=0.1$. This combination is selected over standard $L_2$ or Laplacian regularization to better handle the heavy-tailed noise distributions typical of real-world LiCSAR products while explicitly enforcing the first-order smoothness priors required for geophysical validity.

\begin{figure}[h!]
    \centering
    \includegraphics[width=0.85\linewidth]{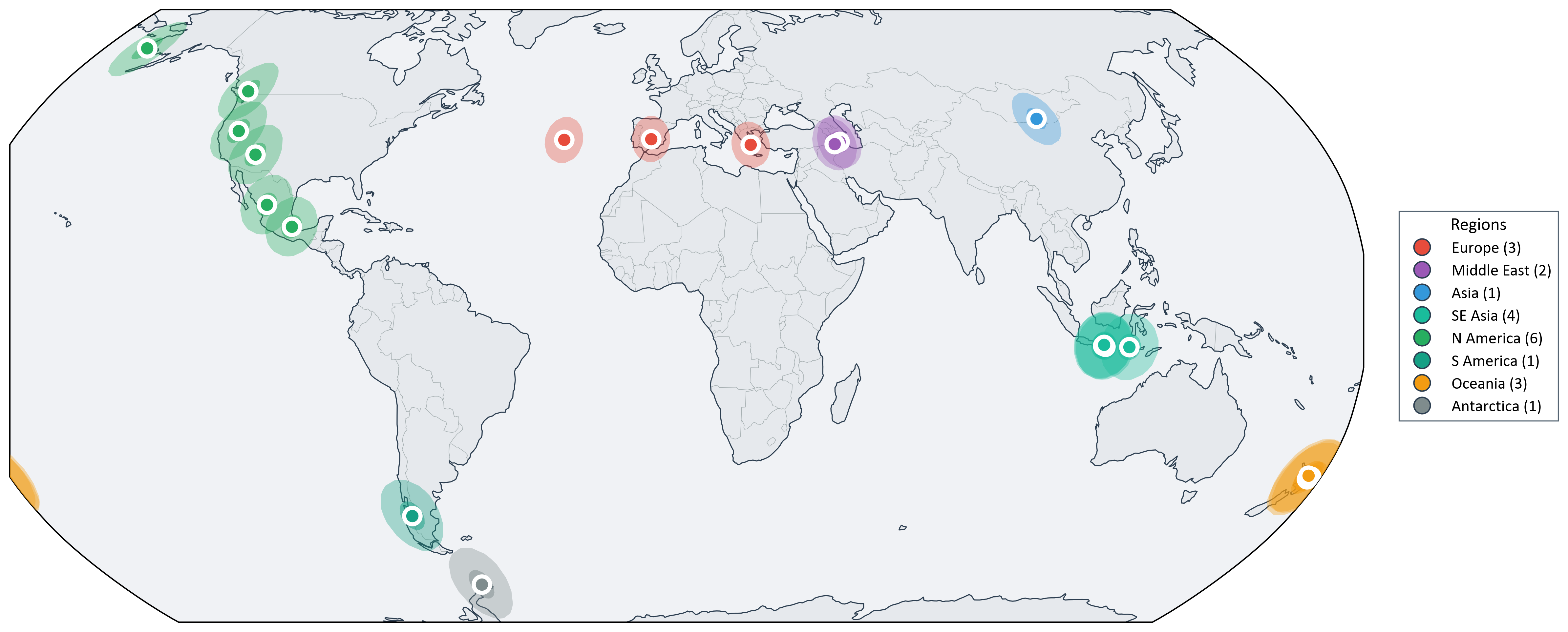}
    \caption{Geographic distribution of 20 LiCSAR frames across 6 continents.}
    \label{fig:map}
\end{figure}

\section{Experimental Framework}

\textbf{Operational Benchmark Construction.} We curate a global InSAR dataset from 350 operational LiCSAR interferograms (2020--2025) \citep{lazecky2020licsar} spanning 20 frames across six continents (Fig. \ref{fig:map}). The dataset encompasses diverse volcanic (e.g., White Island, Pico de Orizaba), tectonic (Middle Gobi, Sahand), and glacio-tectonic (Deception Island) regimes. Each sample integrates wrapped phase, SNAPHU-unwrapped ground truth, coherence ($\gamma \in [0,1]$), and East-North-Up look vectors. From full-frame products, we extract $128 \times 128$ patches (stride = 64) and apply strict quality filters ($\bar{\gamma} > 0.5$, max displacement $> 1$mm), yielding 39,724 high-quality patches (651M pixels).

\textit{Critical Innovation:} To prevent spatial leakage, we implement frame-level stratified splitting, assigning entire geographic regions exclusively to train (14 frames), validation (3 frames), or test (3 frames) sets, ensuring generalization to unseen geographic provinces.

\textbf{Systematic Architectural Ablation.} To isolate the impact of recent computer vision (CV) advancements on geophysical regression, all models utilize an identical 4-level U-Net backbone \citep{ronneberger2015unet} (base channels $C=32$). We evaluate four levels of increasing complexity:
\begin{itemize}
    \setlength\itemsep{0.1em}
    \item \textbf{V-UNet (Vanilla, 7.76M params):} Standard $2 \times (\text{Conv3}\times\text{3} \to \text{BN} \to \text{ReLU})$ blocks with skip connections; our primary baseline for local inductive bias.
    \item \textbf{E-UNet (Enhanced, 8.29M params):} Incorporates Squeeze-Excitation blocks \citep{hu2018senet} after each encoder stage for channel-wise recalibration.
    \item \textbf{A-UNet (Attention, 11.37M params):} Integrates 4-head self-attention at the bottleneck and spatial attention gates at skip connections \citep{schlemper2019attention} for global context.
    \item \textbf{H-UNet (Hybrid, 17.21M params):} Combines SE blocks, MHSA, and an Atrous Spatial Pyramid Pooling (ASPP) \citep{chen2018deeplab} bottleneck to capture multi-scale features (see Appendix~\ref{app:arch_details}).
\end{itemize}

\textbf{Training Protocol.} Models are optimized using AdamW with a OneCycleLR scheduler. To ensure a fair comparison, we perform a validation grid search for each model to determine optimal dropout ($0.0$--$0.2$) and weight decay ($5\times 10^{-5}$--$10^{-4}$), accounting for the increased capacity of larger variants. Attention and Hybrid models use mixed-precision (FP16); Vanilla and Enhanced use full FP32. All models use batch size 32 and early stopping (patience = 100). (see Appendix~\ref{app:training})

\section{Results \& Analysis}

\textbf{Quantitative Performance.} Table \ref{tab:results} summarizes performance across 5,961 geographically held-out patches. The \textbf{Vanilla U-Net consistently achieves the best performance} despite having 32--122\% fewer parameters, revealing a systematic "complexity penalty": attention mechanisms lead to a 25\% $R^2$ drop ($0.834 \to 0.622$) and 51\% RMSE increase. Vanilla U-Net reaches the operational threshold ($<$1cm error) in 88\% of predictions versus only 67.5\% for the Hybrid, confirming convolutional locality better aligns with geophysical regression.

\begin{figure*}[h!] \centering \includegraphics[width=\linewidth]{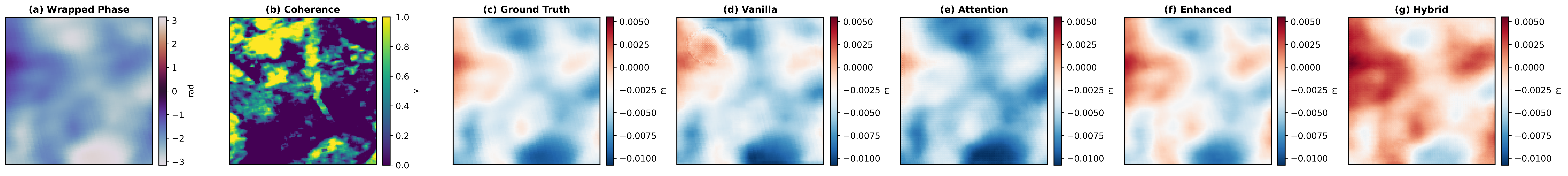} \caption{Representative predictions across test regimes.} \label{fig:qualitative} \end{figure*}

\begin{table}[h!]
\centering
\small
\setlength{\tabcolsep}{8pt} 
\begin{tabular}{lcccccc} 
\toprule
\rowcolor{blue!10} 
\textbf{Model} & \textbf{RMSE (cm)} & \textbf{MAE (cm)} & \textbf{$R^2$} & \textbf{P@1.0 (\%)} & \textbf{95th \%ile (cm)} & \textbf{Params (M)} \\
\midrule
\textbf{Vanilla}  & \textbf{1.009} & \textbf{0.492} & \textbf{0.834} & \textbf{88.0} & \textbf{2.03} & 7.76 \\
Enhanced          & 1.149          & 0.647          & 0.786          & 80.7          & 2.64          & 8.29 \\
Attention         & 1.528          & 0.844          & 0.622          & 75.6          & 3.40          & 11.37 \\
Hybrid            & 1.595          & 1.001          & 0.588          & 67.5          & 3.49          & 17.21 \\
\bottomrule
\end{tabular}
\caption{Test set performance on 5,961 held-out patches. Bold indicates best performance.}
\label{tab:results}
\end{table}

\textbf{Operational Efficiency.} The Vanilla U-Net achieves 2.92ms latency, a 2.5$\times$ speedup over the Hybrid model (Table \ref{tab:efficiency}). The 2.2$\times$ lower memory footprint (29.62MB) is critical for deployment on resource-constrained observatory edge-nodes. While all variants meet the sub-100ms requirement for early warning, Vanilla enables continental-scale monitoring at a fraction of computational cost.

\begin{table}[h!]
\centering
\small
\setlength{\tabcolsep}{8pt}
\begin{tabular}{lcccc}
\toprule
\rowcolor{blue!10}
\textbf{Model} & \textbf{Memory (MB)} & \textbf{FLOPs (G)} & \textbf{Latency (ms)} & \textbf{Params (M)} \\
\midrule
\textbf{Vanilla} & \textbf{29.62} & \textbf{1017.63} & \textbf{2.92 $\pm$ 0.06} & 7.76 \\
Enhanced & 31.61 & 1086.21 & 6.35 $\pm$ 0.07 & 8.29 \\
Attention & 43.38 & 1490.66 & 7.08 $\pm$ 0.07 & 11.37 \\
Hybrid & 65.64 & 2255.24 & 7.13 $\pm$ 0.17 & 17.21 \\
\bottomrule
\end{tabular}
\caption{Operational efficiency profiling (NVIDIA GH200).}
\label{tab:efficiency}
\end{table}

\begin{figure}[h!]
    \centering
    \includegraphics[width=0.45\linewidth]{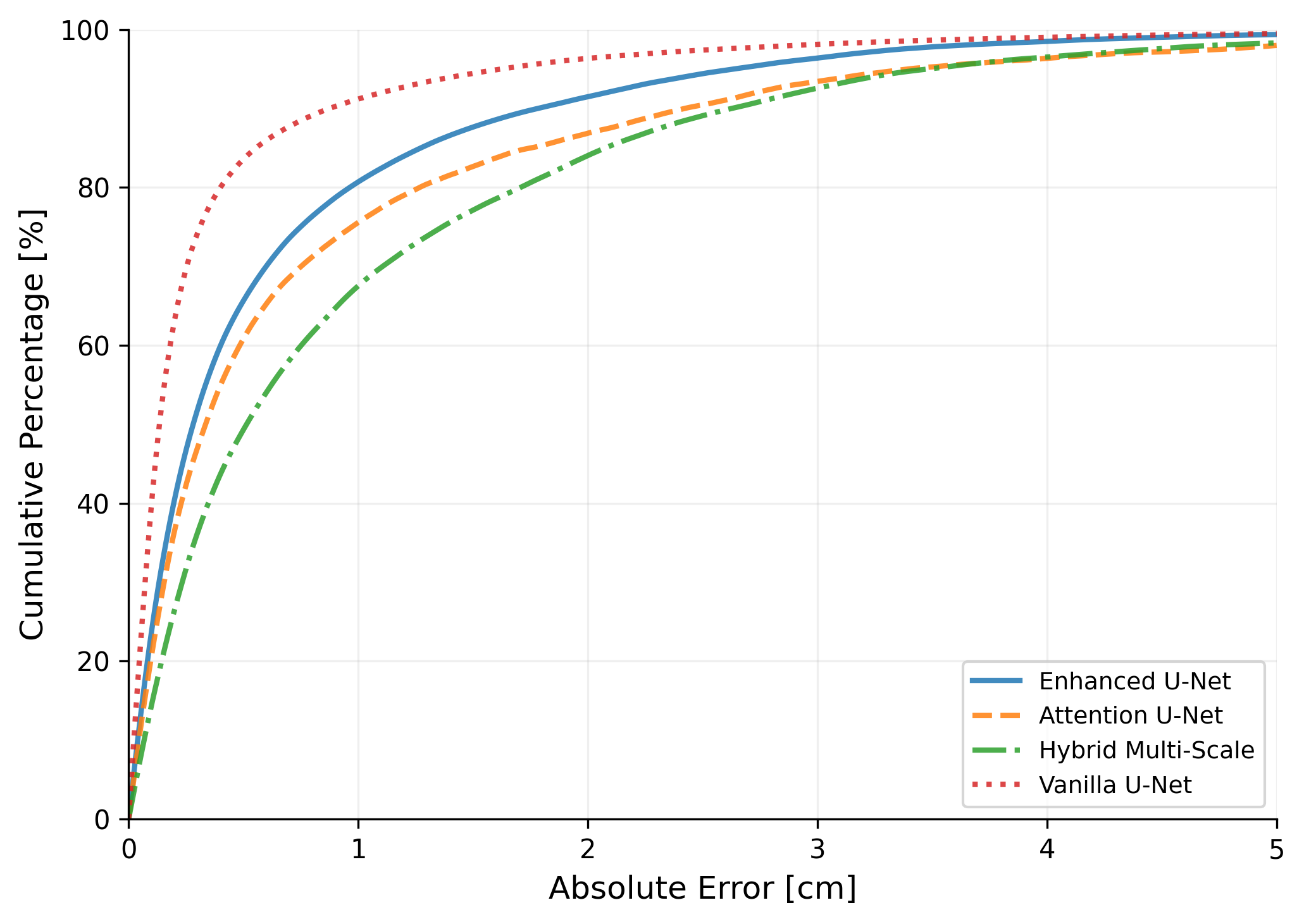}
    \includegraphics[width=0.45\linewidth]{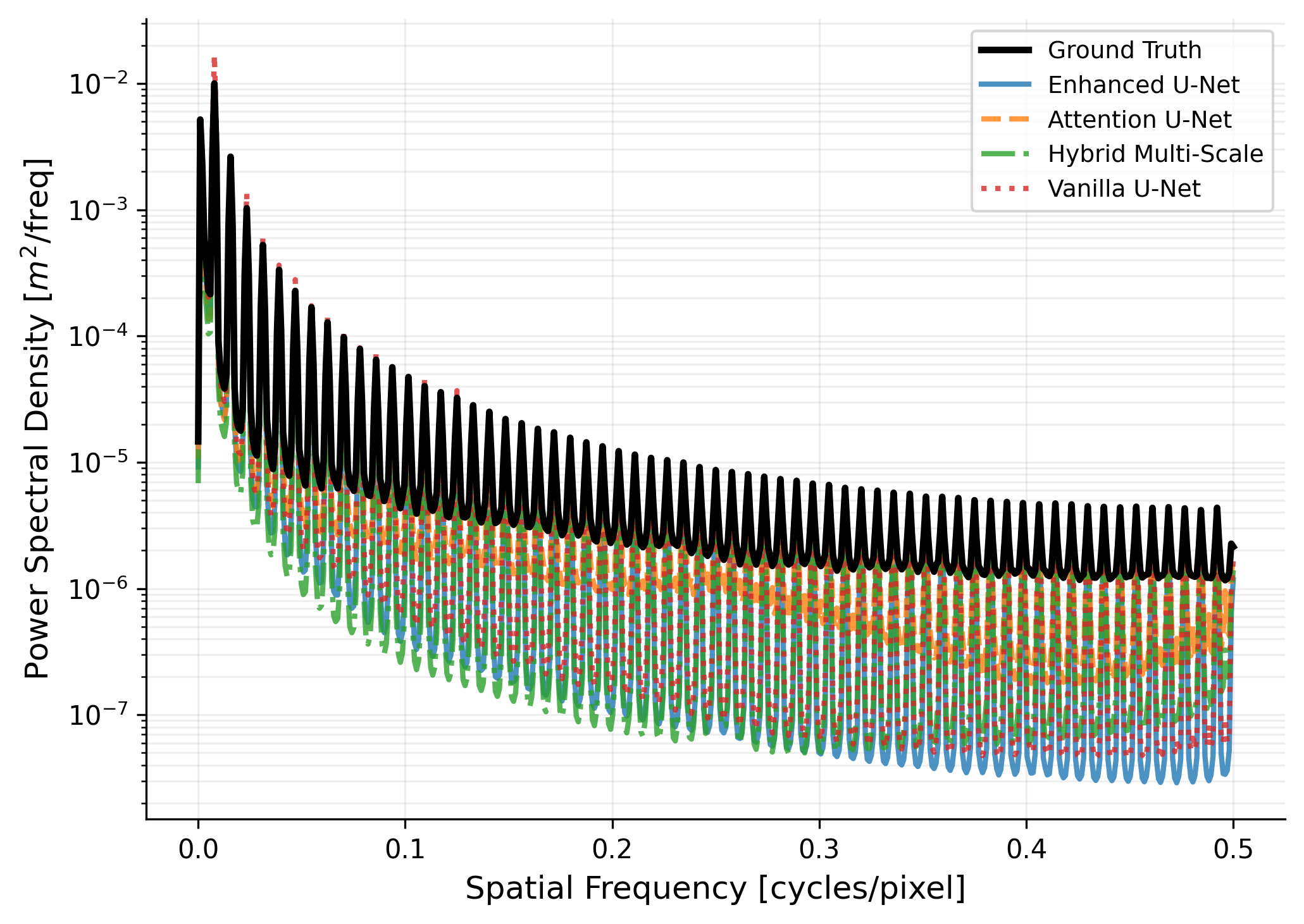}
    \caption{\textbf{(a)} Cumulative error distribution. \textbf{(b)} Power spectral density analysis}
    \label{fig:diagnostics}
\end{figure}

\textbf{Physics-Grounded Failure Analysis.} Power Spectral Density (PSD) analysis (Figure \ref{fig:diagnostics}b) reveals that Vanilla and Enhanced models accurately preserve the ground-truth spectrum. In contrast, Attention and Hybrid models inject spurious high-frequency power at $>0.3$ cycles/pixel. Given that crustal deformation is governed by elasticity, true signals rarely exhibit sub-wavelength variations at the 14m Sentinel-1 scale. Consequently, the high-frequency content produced by complex models represents hallucinated unphysical artifacts rather than legitimate geophysical signal.

\textbf{Root Causes of Failure.} We identify three mechanisms driving this divergence: (1) \textit{Inductive bias mismatch}: Attention mechanisms excel at detecting discrete boundaries in natural images \citep{dosovitskiy2021vit, vaswani2017attention}; however, InSAR displacement is characterized by high spatial autocorrelation, making the global flexibility of attention a liability for continuous fields, disrupting local autocorrelation structure and introducing spurious long-range dependencies. (2) \textit{Capacity-data mismatch}: The 17M-parameter Hybrid models appear to overfit frame-specific atmospheric noise rather than underlying physics, evidenced by degraded generalization to held-out test frames despite strong training performance. (3) \textit{Multi-scale misapplication}: ASPP-driven aggregation introduces aliasing artifacts when regressing the smooth spectral decay characteristic of geophysical deformation. These failure modes are visually confirmed in Figures \ref{fig:qualitative} and \ref{fig:appendix_visuals}: across volcanic, tectonic, and vegetated regimes, Vanilla predictions closely match the smooth gradients of SNAPHU ground truth, whereas attention-based models exhibit unphysical discontinuities and localized artifacts, particularly near patch boundaries and in low-coherence regions.

\section{Discussion \& Conclusion}
\textbf{Design Principles for ML4RS:} (1) Domain ablations are mandatory: ImageNet winners fail when geophysical physics dominates. (2) Match inductive bias to physics: Convolutional locality beats global attention for autocorrelated fields. (3) Validate with physics diagnostics: Spectral analysis reveals violations invisible to RMSE. (4) Simplicity generalizes better: Vanilla models learn physics rather than scene-specific noise. Complexity may suit temporal or multi-modal tasks, but for smooth-field regression, domain physics must guide design.

\textbf{Limitations \& Future Work.} Parameter count differences (7.76M--17.21M) and single split are acknowledged. Future research should explore capacity-matched variants, multi-sensor generalization (ALOS-2/NISAR), and physics-hybrid layers embedding elasticity constraints.

\textbf{Conclusion.} We presented the first systematic architectural ablation for operational InSAR across 20 frames and 651M pixels. Vanilla U-Net outperforms complex variants by 34\% in $R^2$ with 2.5$\times$ faster inference. PSD analysis confirms that architectural complexity injects high-frequency artifacts via inductive bias mismatch. For physics-constrained regression, domain physics, not architectural sophistication, should guide ML4RS design. Less is more.

\clearpage
\bibliography{iclr2026_conference}

@article{chen2001snaphu,
author = {Curtis W. Chen and Howard A. Zebker},
journal = {J. Opt. Soc. Am. A},
number = {2},
pages = {338--351},
publisher = {Optica Publishing Group},
title = {Two-dimensional phase unwrapping with use of statistical models for cost functions in nonlinear optimization},
volume = {18},
month = {Feb},
year = {2001},
url = {https://opg.optica.org/josaa/abstract.cfm?URI=josaa-18-2-338},
doi = {10.1364/JOSAA.18.000338},
}

@InProceedings{chen2018deeplab,
author="Chen, Liang-Chieh
and Zhu, Yukun
and Papandreou, George
and Schroff, Florian
and Adam, Hartwig",
editor="Ferrari, Vittorio
and Hebert, Martial
and Sminchisescu, Cristian
and Weiss, Yair",
title="Encoder-Decoder with Atrous Separable Convolution for Semantic Image Segmentation",
booktitle="Computer Vision -- ECCV 2018",
year="2018",
publisher="Springer International Publishing",
address="Cham",
pages="833--851",
isbn="978-3-030-01234-2"
}

@misc{dosovitskiy2021vit,
      title={An Image is Worth 16x16 Words: Transformers for Image Recognition at Scale}, 
      author={Alexey Dosovitskiy and Lucas Beyer and Alexander Kolesnikov and Dirk Weissenborn and Xiaohua Zhai and Thomas Unterthiner and Mostafa Dehghani and Matthias Minderer and Georg Heigold and Sylvain Gelly and Jakob Uszkoreit and Neil Houlsby},
      year={2021},
      eprint={2010.11929},
      archivePrefix={arXiv},
      primaryClass={cs.CV},
      url={https://arxiv.org/abs/2010.11929}, 
}

@misc{hu2018senet,
      title={Squeeze-and-Excitation Networks}, 
      author={Jie Hu and Li Shen and Samuel Albanie and Gang Sun and Enhua Wu},
      year={2019},
      eprint={1709.01507},
      archivePrefix={arXiv},
      primaryClass={cs.CV},
      url={https://arxiv.org/abs/1709.01507}, 
}

@Article{lazecky2020licsar,
AUTHOR = {Lazecký, Milan and Spaans, Karsten and González, Pablo J. and Maghsoudi, Yasser and Morishita, Yu and Albino, Fabien and Elliott, John and Greenall, Nicholas and Hatton, Emma and Hooper, Andrew and Juncu, Daniel and McDougall, Alistair and Walters, Richard J. and Watson, C. Scott and Weiss, Jonathan R. and Wright, Tim J.},
TITLE = {LiCSAR: An Automatic InSAR Tool for Measuring and Monitoring Tectonic and Volcanic Activity},
JOURNAL = {Remote Sensing},
VOLUME = {12},
YEAR = {2020},
NUMBER = {15},
ARTICLE-NUMBER = {2430},
URL = {https://www.mdpi.com/2072-4292/12/15/2430},
ISSN = {2072-4292},
DOI = {10.3390/rs12152430}
}

@article{reichstein2019deeplearning,
  author  = {Reichstein, Markus and Camps-Valls, Gustau and Stevens, Bjorn and Jung, Martin and Denzler, Joachim and Carvalhais, Nuno and Prabhat},
  title   = {Deep learning and process understanding for data-driven Earth system science},
  journal = {Nature},
  year    = {2019},
  volume  = {566},
  number  = {7743},
  pages   = {195--204},
  month   = {feb},
  doi     = {10.1038/s41586-019-0912-1},
  url     = {https://doi.org/10.1038/s41586-019-0912-1},
  issn    = {1476-4687}
}

@misc{ronneberger2015unet,
      title={U-Net: Convolutional Networks for Biomedical Image Segmentation}, 
      author={Olaf Ronneberger and Philipp Fischer and Thomas Brox},
      year={2015},
      eprint={1505.04597},
      archivePrefix={arXiv},
      primaryClass={cs.CV},
      url={https://arxiv.org/abs/1505.04597}, 
}

@article{schlemper2019attention,
title = {Attention gated networks: Learning to leverage salient regions in medical images},
journal = {Medical Image Analysis},
volume = {53},
pages = {197-207},
year = {2019},
issn = {1361-8415},
doi = {10.1016/j.media.2019.01.012},
url = {https://www.sciencedirect.com/science/article/pii/S1361841518306133},
author = {Jo Schlemper and Ozan Oktay and Michiel Schaap and Mattias Heinrich and Bernhard Kainz and Ben Glocker and Daniel Rueckert}
}

@ARTICLE{spoorthi2019phasenet,
  author={Spoorthi, G. E. and Gorthi, Subrahmanyam and Gorthi, Rama Krishna Sai Subrahmanyam},
  journal={IEEE Signal Processing Letters}, 
  title={PhaseNet: A Deep Convolutional Neural Network for Two-Dimensional Phase Unwrapping}, 
  year={2019},
  volume={26},
  number={1},
  pages={54-58},
  doi={10.1109/LSP.2018.2879184}}

@article{tobler1970first,
    author = {W. R. Tobler},
    title = {A Computer Movie Simulating Urban Growth in the Detroit Region},
    journal = {Economic Geography},
    volume = {46},
    number = {sup1},
    pages = {234--240},
    year = {1970},
    publisher = {Routledge},
    doi = {10.2307/143141},
    URL = {https://www.tandfonline.com/doi/abs/10.2307/143141},
    eprint ={https://www.tandfonline.com/doi/pdf/10.2307/143141}
}

@misc{vaswani2017attention,
      title={Attention Is All You Need}, 
      author={Ashish Vaswani and Noam Shazeer and Niki Parmar and Jakob Uszkoreit and Llion Jones and Aidan N. Gomez and Lukasz Kaiser and Illia Polosukhin},
      year={2023},
      eprint={1706.03762},
      archivePrefix={arXiv},
      primaryClass={cs.CL},
      url={https://arxiv.org/abs/1706.03762}, 
}

@ARTICLE{zhou2021ai_insar,
  author={Zhou, Lifan and Yu, Hanwen and Lan, Yang and xing, Mengdao},
  journal={IEEE Geoscience and Remote Sensing Magazine}, 
  title={Artificial Intelligence In Interferometric Synthetic Aperture Radar Phase Unwrapping: A Review}, 
  year={2021},
  volume={9},
  number={2},
  pages={10-28},
  doi={10.1109/MGRS.2021.3065811}
}
\bibliographystyle{iclr2026_conference}

\newpage

\newpage
\appendix
\section{Architectural Specifications and Design Rationales}
\label{app:arch_details}

\begin{figure*}[h!]
    \centering
    \begin{subfigure}{0.80\textwidth}
        \centering
        \includegraphics[width=\linewidth]{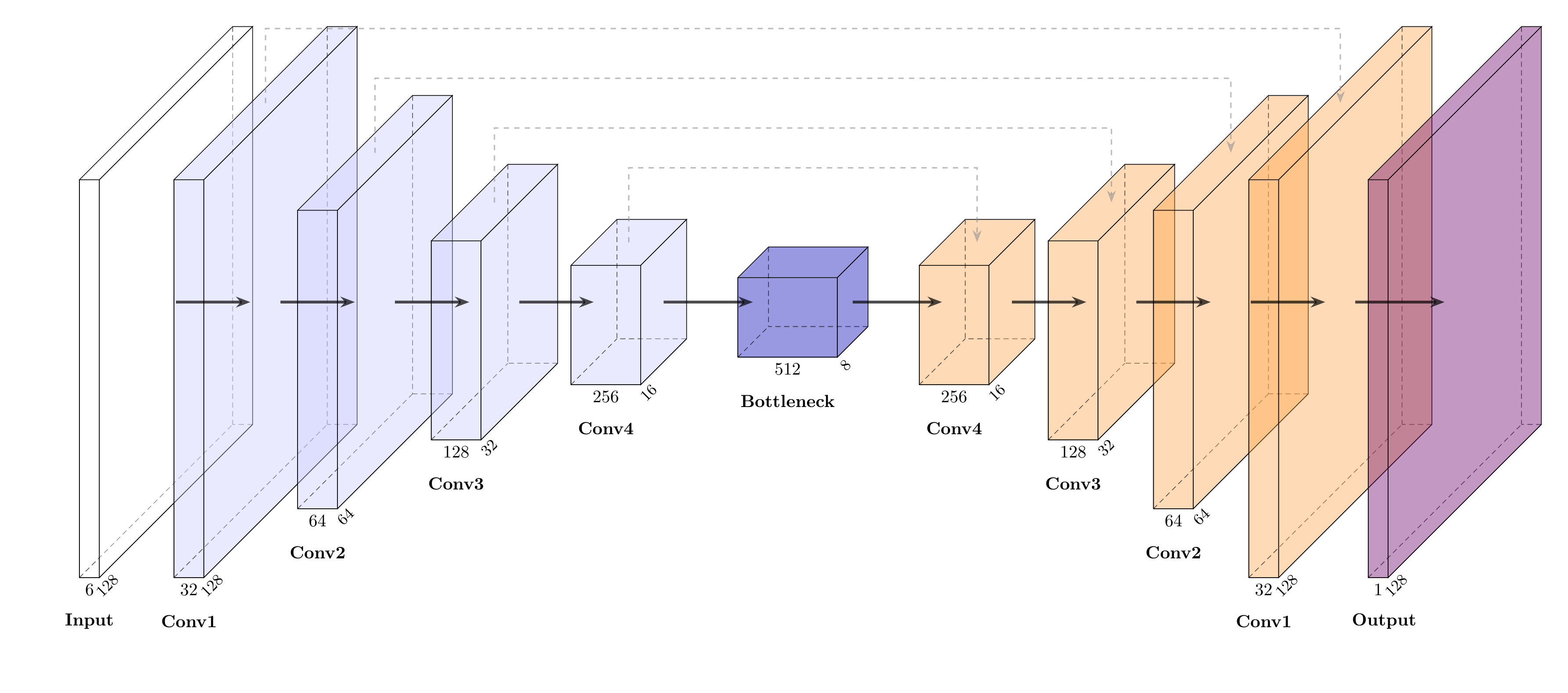}
        \caption{Vanilla U-Net (7.76M params)}
        \label{fig:arch_vanilla}
    \end{subfigure}
    \hfill
    \begin{subfigure}{0.80\textwidth}
        \centering
        \includegraphics[width=\linewidth]{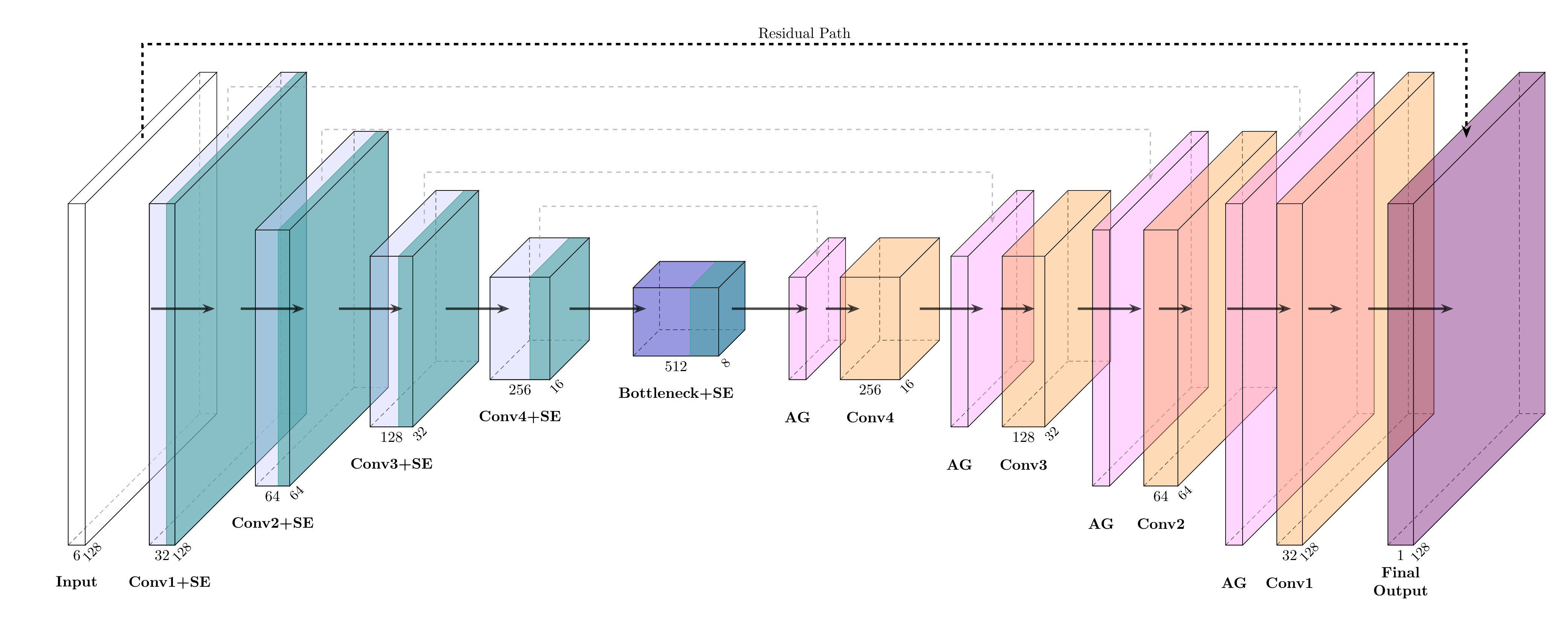}
        \caption{Enhanced U-Net (8.29M params)}
        \label{fig:arch_enhanced}
    \end{subfigure}
    \vspace{1em} 
    \begin{subfigure}{0.80\textwidth}
        \centering
        \includegraphics[width=\linewidth]{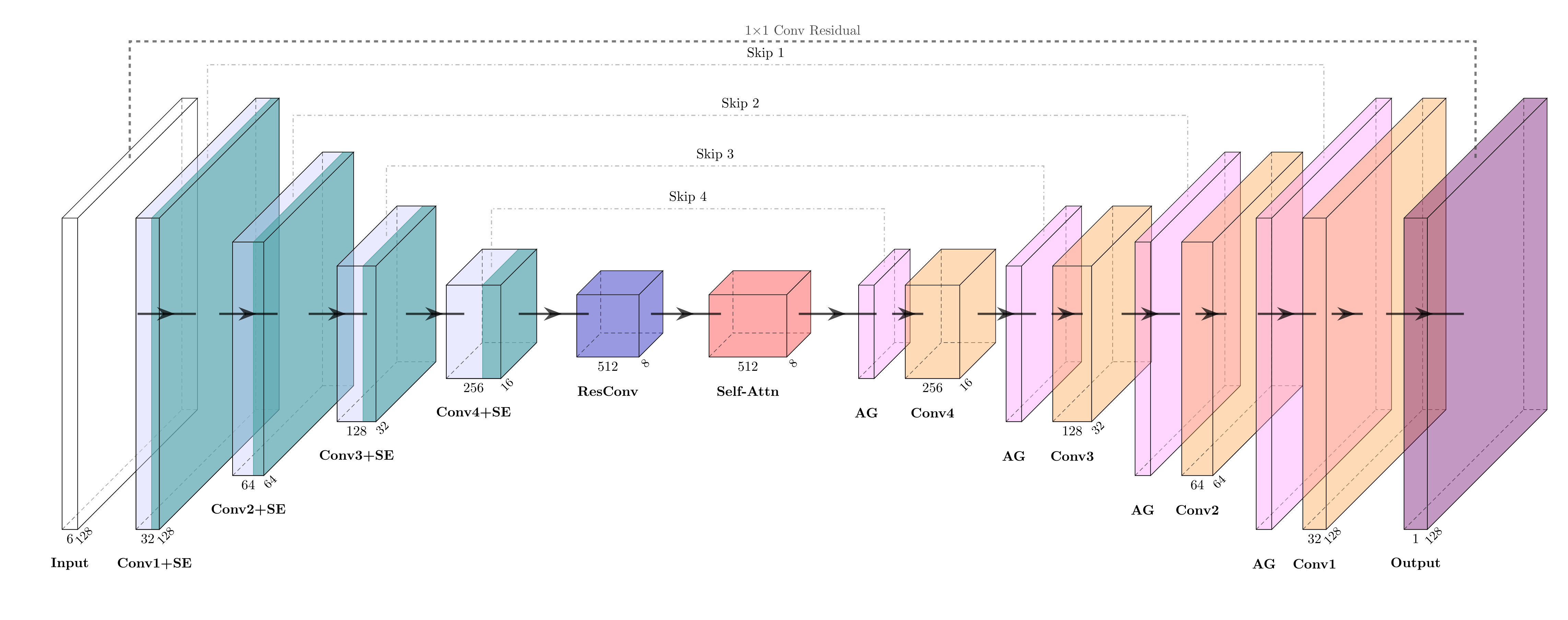}
        \caption{Attention U-Net (11.37M params)}
        \label{fig:arch_attention}
    \end{subfigure}
    \hfill
    \begin{subfigure}{0.80\textwidth}
        \centering
        \includegraphics[width=\linewidth]{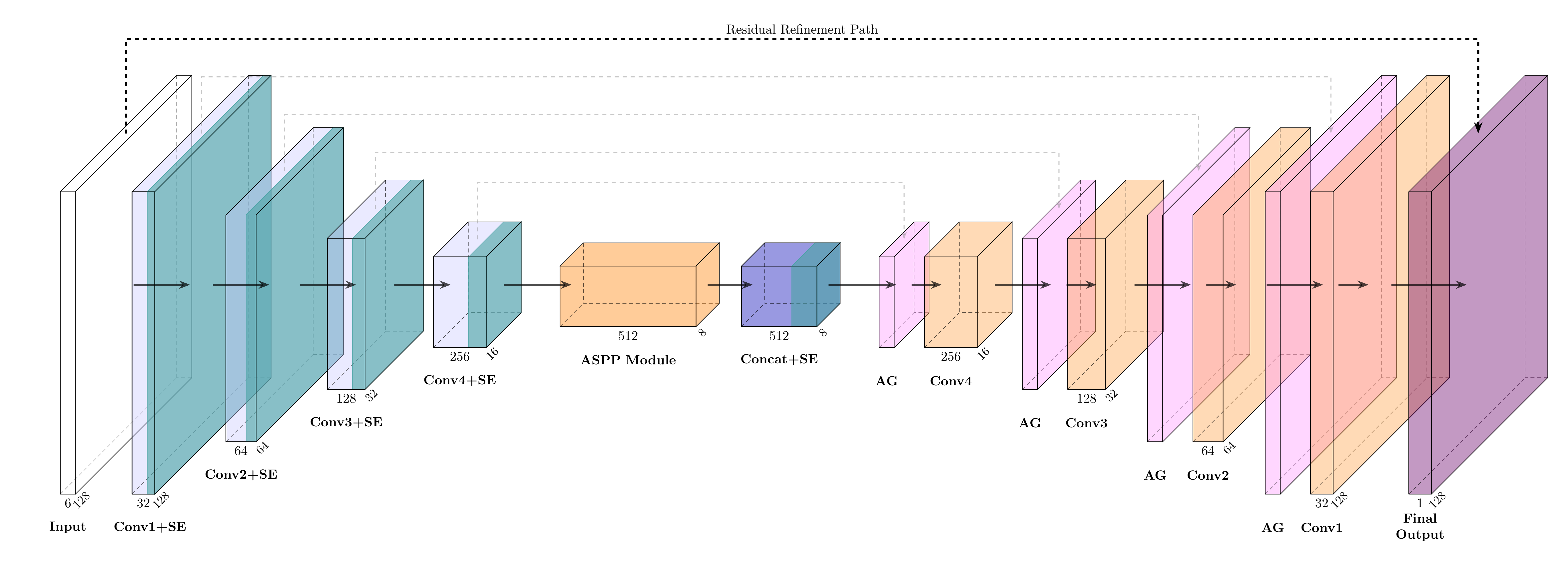}
        \caption{Hybrid Multi-Scale (17.21M params)}
        \label{fig:arch_hybrid}
    \end{subfigure}
    \caption{Detailed architectural variants evaluated in this study}
    \label{fig:architectures}
\end{figure*}

To ensure full reproducibility and provide a technical basis for the "Complexity Penalty" observed in our experiments, we detail the implementation of all four architectural variants (Fig. \ref{fig:architectures}).

\subsection{Variant Specifications}

\textbf{Vanilla U-Net (7.76M parameters):}
Our baseline serves as the minimalist control group, adhering strictly to the original U-Net topology but with modern normalization.
\begin{itemize}
    \item \textbf{Encoder:} 4-level hierarchy. Each level consists of two $3\times3$ convolutions (padding=1).
    \item \textbf{Block Structure:} $\text{Conv3}\times\text{3} \rightarrow \text{BatchNorm} \rightarrow \text{ReLU} \rightarrow \text{Conv3}\times\text{3} \rightarrow \text{BatchNorm} \rightarrow \text{ReLU}$.
    \item \textbf{Downsampling:} $2\times2$ Max Pooling with stride 2.
    \item \textbf{Channel Progression:} $[32, 64, 128, 256, 512]$.
    \item \textbf{Decoder:} Up-convolutions via $\text{ConvTranspose2d}$ followed by concatenation-based skip connections from the corresponding encoder stage.
    \item \textbf{Output Head:} $1\times1$ Convolution mapping to a single-channel displacement map.
\end{itemize}

\textbf{Enhanced U-Net (8.29M parameters):}
This variant investigates whether channel-wise recalibration can improve phase ambiguity resolution in low-coherence regions.
\begin{itemize}
    \item \textbf{Base:} Identical to Vanilla U-Net.
    \item \textbf{Addition:} Squeeze-and-Excitation (SE) blocks integrated after each encoder stage.
    \item \textbf{SE Formulation:} Let $\mathbf{h}$ be the input feature map. The gated scale $\mathbf{s}$ is:
    \begin{equation}
        \mathbf{s} = \sigma(\mathbf{W}_2 \cdot \text{ReLU}(\mathbf{W}_1 \cdot \text{GlobalAvgPool}(\mathbf{h})))
    \end{equation}
    where the final output is $\tilde{\mathbf{h}} = \mathbf{s} \odot \mathbf{h}$.
    \item \textbf{Reduction Ratios:} $r = \{4, 8, 8, 16\}$ for levels 1 through 4, respectively.
\end{itemize}

\textbf{Attention U-Net (11.37M parameters):}
Designed to capture global dependencies and focus on salient deformation regions through spatial gating.
\begin{itemize}
    \item \textbf{Bottleneck:} Multi-head self-attention (4 heads, $d_k = 128$).
    \item \textbf{Self-Attention:} 
    \begin{equation}
        \text{Attention}(\mathbf{Q}, \mathbf{K}, \mathbf{V}) = \text{softmax}\left(\frac{\mathbf{Q}\mathbf{K}^T}{\sqrt{d_k}}\right)\mathbf{V}
    \end{equation}
    \item \textbf{Skip Connections:} Replaced standard concatenation with Gated Spatial Attention. The attention coefficient $\alpha$ is derived from the gating signal $\mathbf{g}$ (lower level) and the skip feature $\mathbf{x}$ (encoder):
    \begin{equation}
        \alpha = \sigma(\psi^T \cdot \text{ReLU}(\mathbf{W}_g \mathbf{g} + \mathbf{W}_x \mathbf{x}))
    \end{equation}
    The gated output is $\tilde{\mathbf{x}} = \alpha \odot \mathbf{x}$.
\end{itemize}

\textbf{Hybrid Multi-Scale U-Net (17.21M parameters):}
Our most complex variant, combining the SE encoder, ASPP bottleneck, and Gated Attention skips to maximize receptive field multi-modality.
\begin{itemize}
    \item \textbf{Bottleneck:} Atrous Spatial Pyramid Pooling (ASPP) utilizing parallel dilated convolutions.
    \item \textbf{ASPP Structure:} 
    \begin{equation}
        \mathbf{f}_{\text{ASPP}} = \text{Concat}(\mathbf{f}_1^{1\times1}, \mathbf{f}_6^{3\times3}, \mathbf{f}_{12}^{3\times3}, \mathbf{f}_{18}^{3\times3}, \mathbf{f}_{\text{global}})
    \end{equation}
    \item \textbf{Effective Receptive Fields:} Rates of $\{3, 13, 25, 37\}$ pixels plus a global average pooling branch.
\end{itemize}

\section{Training Regimes and Hyperparameters}
\label{app:training}

All models were trained using a standardized protocol to ensure the performance differences are attributable solely to architectural capacity.

\subsection{Hyperparameter Configuration}

\begin{table}[h]
\centering
\small
\setlength{\tabcolsep}{8pt}
\begin{tabular}{lcccc}
\toprule
\rowcolor{blue!10}
\textbf{Model} & \textbf{Max LR} & \textbf{Weight Decay} & \textbf{Dropout} & \textbf{Params (M)} \\
\midrule
Vanilla U-Net & $8 \times 10^{-5}$ & $5 \times 10^{-5}$ & 0.0 & 7.76 \\
Enhanced U-Net & $8 \times 10^{-5}$ & $5 \times 10^{-5}$ & 0.10 & 8.29 \\
Attention U-Net & $8 \times 10^{-5}$ & $1 \times 10^{-4}$ & 0.20 & 11.37 \\
Hybrid Multi-Scale & $5 \times 10^{-5}$ & $1 \times 10^{-4}$ & 0.20 & 17.21 \\
\bottomrule
\end{tabular}
\caption{Hyperparameter configurations for all architectural variants.}
\label{tab:hyperparams}
\end{table}

\textbf{Learning Rate Schedule:} We employed the $\text{OneCycleLR}$ scheduler with a 10\% linear warmup phase, followed by a cosine annealing decay to $\eta_{\text{max}}/25$.

\vspace{0.5em}

\noindent\textbf{Optimization:} AdamW with $\beta_1=0.9, \beta_2=0.999$, and $\epsilon=10^{-8}$. We applied gradient clipping with a maximum norm of 1.0 to ensure stability in the Attention and Hybrid models.

\vspace{0.5em}

\noindent\textbf{Mixed-Precision Details:} Attention and Hybrid models were trained with PyTorch's automatic mixed-precision (AMP) with FP16 forward passes and FP32 master weights, with loss scaling handled automatically via \texttt{GradScaler}. BatchNorm layers are kept at FP32 by PyTorch's autocast by default. The Vanilla and Enhanced models were trained in full FP32 precision given their smaller memory footprint.

\vspace{0.5em}

\noindent\textbf{Convergence:} Maximum epochs set to 1000 with an early stopping patience of 100 validation epochs. Typical convergence occurred between 300--500 epochs.

\section{Data Preprocessing and Quality Control}
\label{app:data}

\textbf{Patch Extraction:} Input interferograms (2000--3000 pixels squared) were decomposed into $128 \times 128$ patches with a 64-pixel stride (50\% overlap) to preserve spatial continuity.

\vspace{0.5em}

\noindent\textbf{Quality Filtering:} To prevent model bias from decorrelated noise, we applied the following strict inclusion criteria:
\begin{itemize}
    \item \textbf{Mean Coherence:} $\bar{\gamma} > 0.5$ (eliminates water bodies and dense vegetation).
    \item \textbf{Signal Threshold:} Maximum displacement $> 1$ mm.
    \item \textbf{Data Integrity:} $> 95\%$ valid pixels per patch.
\end{itemize}
\textbf{Normalization:} Channel-wise training statistics were computed once: $\mathbf{x}_{\text{norm}} = (\mathbf{x} - \mu_{\text{train}}) / (\sigma_{\text{train}} + 10^{-8})$.

\section{Computational Resources and Efficiency}

\textbf{Hardware:} All experiments were conducted on an NVIDIA GH200 GPU (120GB VRAM). 
\textbf{Training Time:} 
\begin{itemize}
    \item \textbf{Vanilla:} $\sim$8 hours.
    \item \textbf{Enhanced:} $\sim$10 hours.
    \item \textbf{Attention:} $\sim$11 hours.
    \item \textbf{Hybrid:} $\sim$16 hours.
\end{itemize}
Total computational budget: $\sim$100 GPU-hours including search and validation runs.

\section{Extended Visual Results}
\label{app:visuals}

\begin{figure*}[ht]
    \centering
    \begin{subfigure}{\linewidth}
        \centering
        \includegraphics[width=\linewidth]{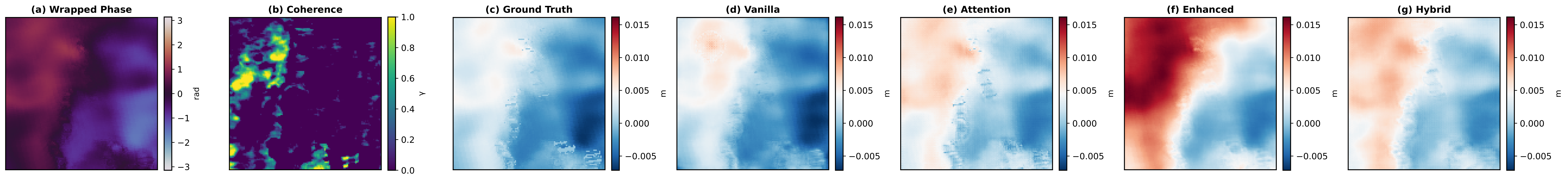}
    \end{subfigure}
    
    \vspace{1.5em}
    
    \begin{subfigure}{\linewidth}
        \centering
        \includegraphics[width=\linewidth]{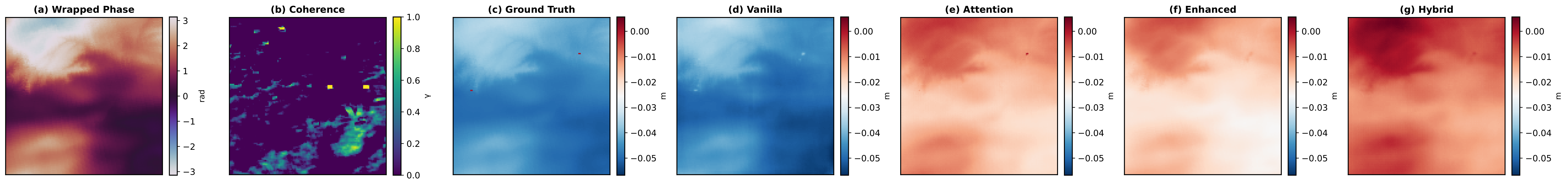}
    \end{subfigure}
    
    \vspace{1.5em}
    
    \begin{subfigure}{\linewidth}
        \centering
        \includegraphics[width=\linewidth]{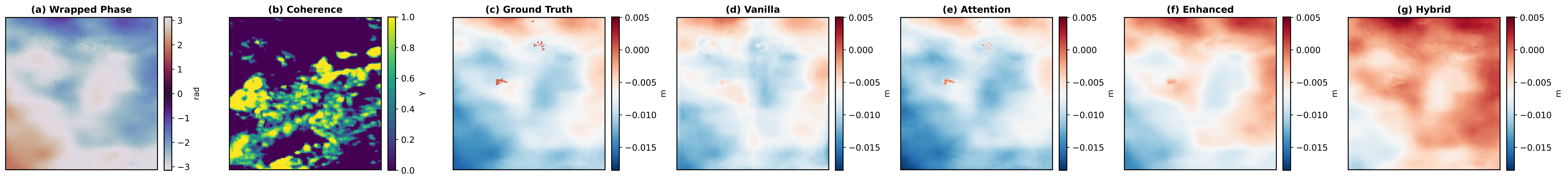}
    \end{subfigure}
    
    \vspace{1.5em}
    
    \begin{subfigure}{\linewidth}
        \centering
        \includegraphics[width=\linewidth]{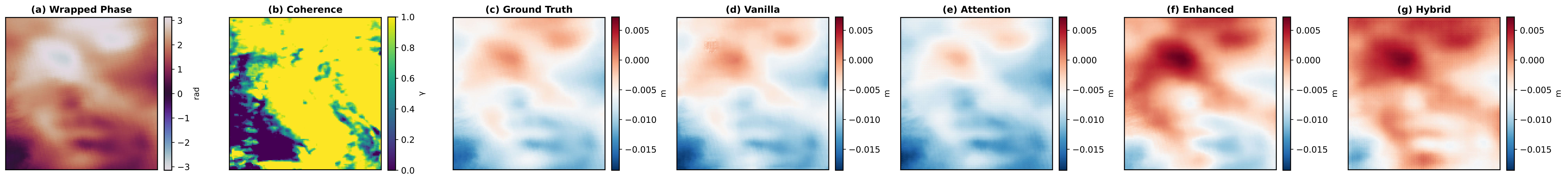}
    \end{subfigure}
    
    \caption{Visual comparison of phase unwrapping results. Each grid presents: (a) Wrapped Phase, (b) Coherence, (c) Ground Truth, (d) Vanilla, (e) Attention, (f) Enhanced, and (g) Hybrid predictions.}
    \label{fig:appendix_visuals}
\end{figure*}

Figure \ref{fig:appendix_visuals} provides a systematic visual comparison across four distinct geographic and tectonic settings, highlighting the robustness of the Vanilla baseline compared to the artifact-prone Hybrid model.

\end{document}